\documentclass[conference]{IEEEtran}
\IEEEoverridecommandlockouts
\usepackage{cite}
\usepackage{amsmath,amssymb,amsfonts}
\usepackage{algorithmic}
\usepackage{graphicx}
\usepackage{textcomp}
\usepackage{xcolor}
\def\BibTeX{{\rm B\kern-.05em{\sc i\kern-.025em b}\kern-.08em
    T\kern-.1667em\lower.7ex\hbox{E}\kern-.125emX}}

\usepackage{graphicx}
\usepackage{multirow}
\usepackage{booktabs}
\usepackage{amsmath,amssymb} 
\usepackage{amsfonts,bm}

\newcommand{\E}{\mathbb{E}}                   
\newcommand{\z}{{\rm\bf z}}                   
\newcommand{\w}{{\rm\bf w}}                   
\newcommand{\Z}{\mathcal{Z}}                  
\newcommand{\W}{\mathcal{W}}                  
\newcommand{\x}{{\rm\bf x}}                   
\newcommand{\X}{\mathcal{X}}                  
\newcommand{\n}{{\rm\bf n}}                   
\newcommand{\Loss}{\mathcal{L}}               
\newcommand{\y}{{\rm\bf y}}       
\newcommand{\A}{{\rm\bf A}}       
\newcommand{\bias}{\rm\bf b}    
\newcommand{\R}{\mathbb{R}}       

\begin{document}

\title{3D Cartoon Face Generation with Controllable Expressions from a Single GAN Image}

\author{
\IEEEauthorblockN{Hao Wang\textsuperscript{1}, Wenhao Shen\textsuperscript{1}, Guosheng Lin\textsuperscript{1}, Steven Hoi\textsuperscript{2}, Chunyan Miao\textsuperscript{1$\dagger$}}
\IEEEauthorblockA{\textsuperscript{1}{Nanyang Technological University, Singapore} \\
\textsuperscript{2}{Singapore Management University, Singapore} 
\\
wanghao-tech@outlook.com
}
}

\maketitle

\renewcommand{\thefootnote}{}
\footnotetext{$\dagger$: Corresponding author.} 

\begin{abstract}
In this paper, we investigate an open research task of generating 3D cartoon face shapes from single 2D GAN generated human faces and without 3D supervision, where we can also manipulate the facial expressions of the 3D shapes. To this end, we discover the semantic meanings of StyleGAN latent space, such that we are able to produce face images of various expressions, poses, and lighting conditions by controlling the latent codes. Specifically, we first finetune the pretrained StyleGAN face model on the cartoon datasets. By feeding the same latent codes to face and cartoon generation models, we aim to realize the translation from 2D human face images to cartoon styled avatars. We then discover semantic directions of the GAN latent space, in an attempt to change the facial expressions while preserving the original identity. As we do not have any 3D annotations for cartoon faces, we manipulate the latent codes to generate images with different poses and lighting conditions, such that we can reconstruct the 3D cartoon face shapes. We validate the efficacy of our method on three cartoon datasets qualitatively and quantitatively.
\end{abstract}


\section{Introduction}
\label{sec:intro}

Translating 2D human faces into 3D personalized cartoon avatars remains challenging, since most of the existing works \cite{kanazawa2018learning,tulsiani2018multi} require rich multi-view 2D information or depth annotations, which are difficult to obtain massively. Moreover, to realize the synchronization between the 2D human faces and the 3D stylized avatars, we need to not only do the style transfer but also enable the 3D avatars to share the same expressions as the original human faces.

\begin{figure}
\begin{center}
\includegraphics[width=0.35\textwidth]{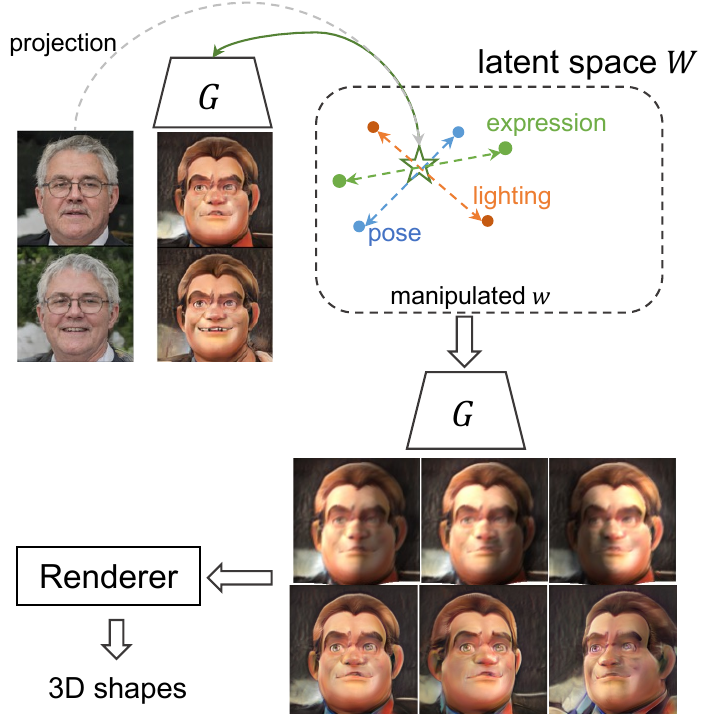}
\end{center}
   \caption{Overview of our proposed pipeline, where we train StyleGAN models for human faces and cartoon datasets respectively. Given a single GAN generated human face image, we first discover its corresponding latent codes $\w$ in the latent space $\W$, where we input $\w$ to the cartoon generator $G$ and generate the cartoon faces. We then aim to uncover the semantic directions of $\w$, by which we can manipulate the facial expressions, poses and lighting conditions of the generated images. The manipulated images are fed into neural renderer for 3D reconstruction.}
\label{fig:demo}
\end{figure}

In this paper, we demonstrate a new application, where we aim to generate 3D cartoon avatars based on single GAN generated human faces and without any 3D supervision. Notably, we do not need any annotations of the multi-view face images or the depth information. Specifically, our proposed task consists of two branches: one is to generate the stylized cartoon images given the GAN generated human face images, in which we can manipulate the facial expressions; another one is to reconstruct the 3D shapes from single 2D images only. The main challenge of our proposed task is to realize the consistency between the 2D human faces, the 2D stylized cartoon faces, and the 3D reconstructed shapes. 

Previous works \cite{karras2019style,Karras2019stylegan2,shen2021closed,wang2021cycle} have demonstrated that the disentanglement of StyleGAN latent space $\W$. That means we can manipulate the latent codes in space $\W$ to change one semantic attribute while preserving other attributes. Therefore, the StyleGAN \cite{Karras2019stylegan2} model can produce images of the same person, but with various facial expressions, poses and lighting conditions, as is shown in Figure \ref{fig:demo}.

As the preliminary of our proposed method, we assume the given human face images are generated from a StyleGAN model pretrained on FFHQ \cite{karras2019style} dataset. If the input is a real-world face image, GAN inversion methods \cite{Karras2019stylegan2,zhu2016generative,bau2020semantic,zhu2020indomain} can be used to project the given face image back to the latent space $\W$ through the face generator, so that we can obtain the corresponding latent code.
We then do the image-to-image translation to add cartoon style onto the original face images. Specifically, we finetune the pretrained FFHQ model to train a cartoon face generation model, in which we feed the same latent codes to both the FFHQ and cartoon generation models, and regularize the training through the output of the intermediate layers. This allows these two generation models to share similar latent space $\W$, making the cartoon and human face images generated from the same latent codes have similar semantic attributes. 

To manipulate the StyleGAN latent codes and generate images with various facial expressions, we use closed-form factorization proposed in SeFa \cite{shen2021closed} to give the manipulated offsets for the latent codes. However, we find the initial offsets discovered by SeFa would also make some undesired semantic concepts changed, such as the person face identity. To resolve this issue, we further optimize the initial offsets by the face identity loss and the low-level feature loss. As a result, we obtain StyleGAN latent codes with expressions changed only. Based on the obtained latent codes, we then discover its variations for different poses and lighting conditions, which are used as the pseudo samples to reconstruct the 3D shapes. During the 3D reconstruction process, we apply regularization on the face identity and the symmetry structures, to alleviate the issues of texture and geometry distortion.

Our contributions can be summarized as follows: 
\begin{itemize}
    \item We propose a novel framework to generate 3D cartoon face shapes given single GAN generated human face images, which does not require any 3D supervision.
    \item We propose to generate 3D cartoon shapes with controllable expressions, where we optimize the manipulated latent code offsets, such that we can modify the facial expressions and keep other semantic attributes unchanged.
    \item We show promising 3D cartoon generation results both qualitatively and quantitatively, in which our model can generate high-quality 3D cartoon face shapes, with controllable expressions, poses, and lighting conditions.
\end{itemize}

\section{Related Work}

\subsection{Image manipulation}
Image manipulation \cite{ak2022learning,karaouglu2021self} aims to change the semantic attributes of the given images. Some works \cite{dong2017semantic,wang2021cycle} adopt the textual descriptions to manipulate images. Specifically, Wang et al. \cite{wang2021cycle} use GAN inversion \cite{zhu2020indomain} to project the images back to the StyleGAN \cite{karras2019style} latent space, and use paired image-text data to align the text embeddings with the GAN latent space, so that the images can be manipulated by the text features. To unsupervisedly discover meaningful latent directions of a pretrained GAN model, GANSpace \cite{harkonen2020ganspace} and SeFa \cite{shen2021closed} use Principal Component Analysis (PCA) to analyse the latent space.  
However, the discovered semantic directions by these methods are still coupled \cite{pan20202d} with other semantic concepts. In our work, we propose to optimize the directions found in SeFa, to make the manipulated face images change expressions only, while keeping other semantic attributes unchanged.

\subsection{Unsupervised 3D shape learning}
Since the 3D shape reconstruction requires images of consistent multiple views and lighting,
recent works \cite{pan20202d,shi2021lifting,kanazawa2018learning,tulsiani2018multi,wu2020unsupervised,tu20203d,liu2020dlgan} attempt to uncover extra cues to guide the learning process. Kanazawa et al. \cite{kanazawa2018learning} use an image collection under the same category as supervision to learn the reconstruction model. Wu et al. \cite{wu2020unsupervised} takes an autoencoding pipeline, where they infer the depth, albedo, viewpoint and lighting from a single image, and use the reconstruction loss to supervise the training.
\cite{pan20202d,shi2021lifting,zhang2020image} aim to manipulate the latent codes of StyleGAN \cite{karras2019style,Karras2019stylegan2} to generate synthetic data for 3D shape learning. However, Zhang et al. \cite{zhang2020image} require the manual annotations for different views. Tu et al. \cite{tu20203d} take the sparse 2D facial landmark heatmaps as additional information to improve 3D face model reconstruction learning.
Shi et al. \cite{shi2021lifting} propose to train a 3D generator to disentangle the latent codes into 3D components, which are used as the input for the renderer. However, they \cite{shi2021lifting} try to infer the depth information without any constraints from the latent codes, this may result in the estimated depth not being precise. Pan et al. \cite{pan20202d} adopts an ellipsoid shape as the shape prior, giving better estimations on the face depth.

\begin{figure*}
\begin{center}
\includegraphics[width=\textwidth]{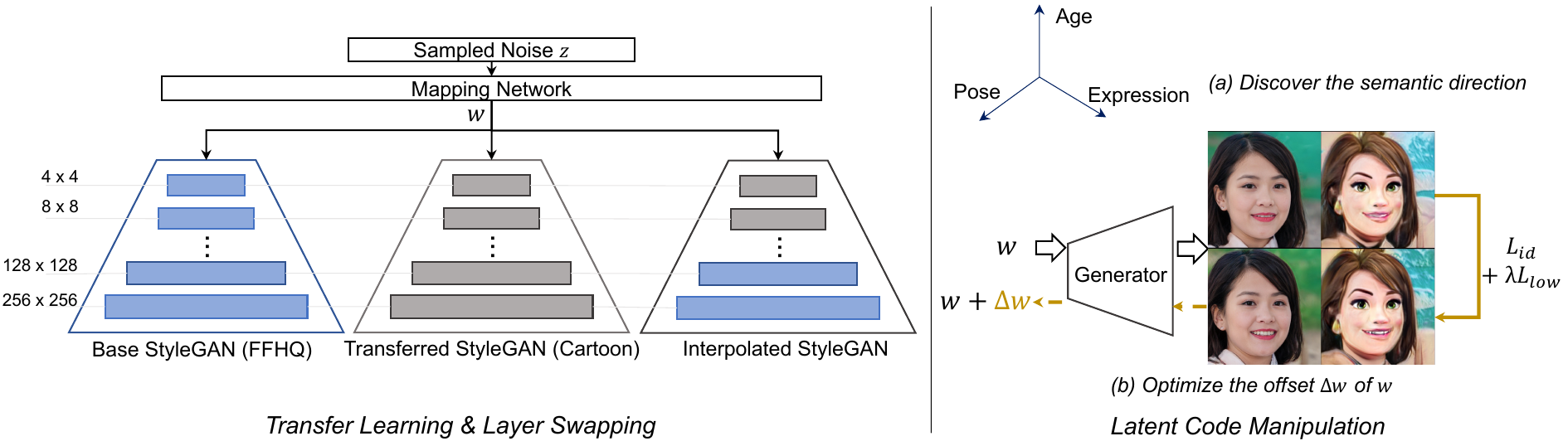}
\end{center}
   \caption{The demonstration of the 2D cartoon face generation model training and latent code manipulation process. 
   We first finetune the pretrained FFHQ StyleGAN model on the cartoon dataset. It is notable that we feed the same latent codes into the cartoon generator as that to the human face image generator. We then interpolate the transferred model, such that we can generate photo-realistic cartoon images. In the latent code manipulation phase, we first discover the semantic directions in the latent space $\W$ of the trained StyleGAN model. We then optimize the offset $\Delta \w$ to the original latent code $\w$ with the identity loss $\Loss_{id}$ and low-level feature regularization loss $\Loss_{low}$. }
\label{fig:framework}
\end{figure*}

\section{Method}
Our proposed framework for 3D cartoon face generation from single 2D GAN generated images is shown in Figure \ref{fig:framework} and \ref{fig:3d}. The whole pipeline can be summarized as a three-stage learning scheme:
\begin{itemize}
   \item \textbf{Stage 1.} We first finetune the pretrained FFHQ \cite{karras2019style} StyleGAN model with the cartoon dataset. Then we interpolate the cartoon generation model weights to generate 2D photo-realistic cartoon faces. 
   \item \textbf{Stage 2.} We aim to discover the semantic directions of the StyleGAN latent space $\W$. To manipulate one semantic concept only of the generated images, while keeping the face identity unchanged, we propose to optimize the offsets $\Delta \w$ with the identity loss $\Loss_{id}$ and the low-level feature regularization loss $\Loss_{low}$. 
   \item \textbf{Stage 3.} Since StyleGAN can generate images of various poses and lighting conditions, we use it to give pseudo samples to reconstruct the 3D shapes of cartoon faces, which are generated based on the discovered original and manipulated latent codes. The face identity and symmetry losses are utilized to improve the reconstructed results. 
\end{itemize}

In the following sections, we give more technical details.

\subsection{StyleGAN preliminaries}
Prevailing generative adversarial networks (GANs) take the randomly sampled noise codes $\z$ as the input and produce fake images, which can be denoted as $G(\cdot): \Z\rightarrow\X$. Karras et al. \cite{karras2019style,Karras2019stylegan2} propose to map random noise $\z \in \Z$ to the latent code $\w \in \W$ with the Multi-Layer Perceptron (MLP). The dimensions are  $\z \in \R^{512}$ and $\w \in \R^{512}$ respectively. Recent works \cite{karras2019style,Karras2019stylegan2,shen2021closed,wang2021cycle} demonstrate that the style-based generator can produce high-fidelity images, and the learned latent space $\W$ is disentangled with respect to various semantic attributes. This means we may manipulate the latent codes $\w$ and generate face images with different facial expressions, poses and lighting conditions, that are useful for the 3D shape reconstruction.

\subsection{2D cartoon generation model}
\label{exp:cartoon}
As shown in the right column of Figure \ref{fig:framework}, we propose to use transfer learning to train a StyleGAN model on the cartoon dataset. The reasons that we adopt the transfer learning technique have two folds: 1) the available samples are limited in most cartoon datasets, meaning training a good cartoon generation model from scratch is challenging, and 2) we want the generated cartoon images to keep similar with the original face images, i.e. sharing similar semantic contents with the given human face images, such as gender and facial expressions. We observe finetuning a well-trained face StyleGAN model allows us to satisfy the above requirements.

To be specific, we take the StyleGAN model pretrained on FFHQ \cite{karras2019style} as the base model, and finetune it on the cartoon datasets. 
During this process, we first aim to generate images having certain cartoon styles, which is achieved by the adversarial training method on the transferred model. 
The adversarial loss can be denoted as:
\begin{align}
    \Loss_{adv} &= \underset{{\x \sim p_{data}}}\E\left[\log D_{trans}(\x)\right] \nonumber \\
&+\underset{{\z \sim p_{\z}}}\E\left[\log \Big(1-D_{trans}\big(G_{trans}(\z)\big)\Big)\right],
\end{align}
where $\x$ denotes the generated cartoon images, $D_{trans}$ is the discriminator to discriminate if the generated images are real or fake. $G_{base}$ and $G_{trans}$ refer to the base and transferred generation models respectively. $p_{data}$ is the distribution of the given cartoon image dataset. $p_{\z}$ is the random noise distribution, which is Gaussian distribution.

Based on our observations, if we only use $\Loss_{adv}$ during the finetuning process, the generated cartoon images would completely follow the cartoon dataset distribution. As a consequence, the cartoon and the human face images generated from the same StyleGAN latent codes would fail to share the same semantic information, such as gender. 
To address this issue, we also apply the structure loss \cite{back2021fine} between the base model and the transferred model. 
Technically, since StyleGAN takes the progressive generation strategy, we can obtain the generated images from various resolutions. In the structure loss, we enforce the low-resolution layer outputs of the base model and that of the transferred model to be similar. This practice manages the trade-off between the cartoon image generation quality and the semantics similarity between the cartoon and human faces. 
The structure loss can be formulated as:
\begin{equation}
    \Loss_{str} = \frac{1}{k}\sum_{i=1}^{k}{||G^{i}_{base}(\z)- G^{i}_{trans}(\z)||^2}, 
\end{equation}
where $i$ denotes the index of the StyleGAN block, $k$ represents we apply $\Loss_{str}$ on the first $k$ blocks.  

The overall training objective of the cartoon generation model is
\begin{equation}
    \Loss_{gen} = \Loss_{adv} + \Loss_{str}.
\end{equation}
It is notable that the weights of $G_{base}$ are fixed during the finetuning process.

During the inference phase, we further adopt the model interpolation \cite{pinkney2020resolution} technique to enable the interpolated StyleGAN model to produce more photo-realistic cartoon images. When we want to generate cartoon styled images that have similar semantics to the GAN generated human face images, we only need to feed the same latent codes $\w$ into the cartoon generator as that to the human face image generator. 

Our proposed method is an efficient way to realize the translation to the high-fidelity cartoon avatars.

\begin{figure*}
\begin{center}
\includegraphics[width=0.7\textwidth]{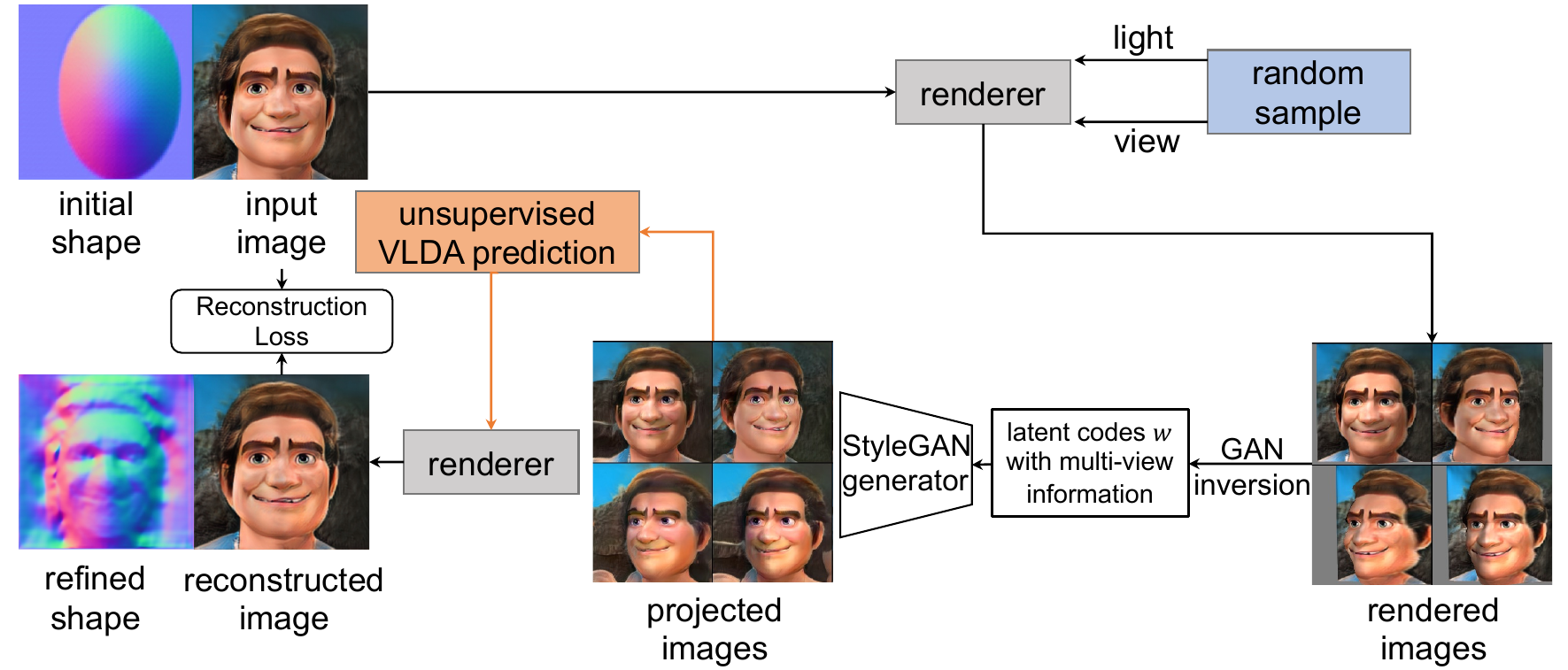}
\vspace{-10pt}
\end{center}
   \caption{The demonstration of 3D cartoon shape reconstruction process. VLDA stands for the viewpoint, lighting conditions, depth and albedo respectively. Given an input image, we feed it into the renderer with the initial shape prior. We randomly sample lighting conditions and viewpoints, generating rendered images from various viewpoints and lighting. These rendered results from the 3D shapes are further projected back to the latent space $\W$ of StyleGAN. This gives better quality for the projected images, which are used to refine initial shapes. The reconstruction loss is applied on the input and reconstructed images.}
\label{fig:3d}
\end{figure*}

\subsection{Latent code manipulation} \label{sec:latent}
To control the semantic attributes of generated avatars, such as the facial expressions, we aim to discover the semantic directions of the StyleGAN latent space $\W$. Shen et al. \cite{shen2021closed} propose to use closed-form factorization to find the meaningful directions. To be specific, Shen et al. represent the transformation from the latent code $\z$ to $\w$ as
\begin{align}
  G_1(\z) \triangleq \y = \A\z + \bias,
\end{align}
where $\A$ and $\bias$ denote the weight and bias. Further, the manipulated latent codes are 
\begin{align}
  G_1(\z') = G_1(\z + \alpha\n) = \y + \alpha\A\n,
\end{align}
where $\alpha$ indicates the manipulation intensity, and $\n$ denotes a certain direction in the latent space to represent a semantic attribute \cite{shen2021closed,goetschalckx2019ganalyze,shen2020interpreting,yang2021semantic}. As discussed in \cite{shen2021closed}, the weight parameter $\A$ of the mapping network may contain the essential knowledge of the image variation. That means the latent semantic directions can be discovered by decomposing $\A$. To this end, we solve the following optimization problem
\begin{align}
  \n^* = \mathop{\arg\max}_{\{\n\in\R^d:\ \n^T\n = 1\}} ||\A\n||_2^2, 
\end{align}
where $||\cdot||_2$ indicates the $l_2$ norm. 

However, we observe that manipulating the latent code with the discovered semantic direction $\n^*$ \cite{shen2021closed} usually does not change one attribute only, it may also affect some other irrelevant semantic concepts, e.g. the face identity. To resolve this issue of \cite{shen2021closed}, we propose to further optimize the offset $\Delta \w$ to the latent code. Technically, we use $\alpha\n^*$ as the initial offset. We then adopt a pre-trained face recognition network \cite{shi2021lifting} $f(\cdot)$ to regularize the face identities, which can be denoted as
\begin{align} \label{eq:id}
  \Loss_{id} &= ||f\big(G_{base}(\w)\big)- f\big(G_{base}(\w+\Delta \w)\big)||^2. 
\end{align}
Note that here we use the base StyleGAN model $G_{base}$ for the optimization process, since there is no available cartoon face recognition network, and the transferred cartoon StyleGAN model has been trained to have similar latent space $\W$ to $G_{base}$.

To avoid getting trivial optimized results, where $\Delta \w \rightarrow 0$, we propose to add another constraint to maximize the low-level features between the original and the manipulated face images, such that the manipulated images could have obvious texture differences to the original images. The low-level features capture more on the minor details and boundary information of faces. With a segmentation model \cite{yu2018bisenet}, we feed the face areas only into the feature extractor $f_{low}(\cdot)$, which mitigates the background noise. The low-level feature regularization loss is
\begin{align}
  \Loss_{low} = ||f_{low}\big(G_{base}(\w)\big)- f_{low}\big(G_{base}(\w+\Delta \w)\big)||^2.
\end{align}
The optimization objective can be formulated as
\begin{align}
  \Loss_{opt} &= \Loss_{id} - \lambda_{low} \Loss_{low}. \label{eq:lm}
\end{align}
where $\lambda_{low}$ is the trade-off hyper-parameter.

\subsection{3D shape reconstruction}
In Figure \ref{fig:3d}, we present the process of 3D reconstruction from single 2D cartoon images. To reconstruct the 3D avatar shape from a single image, we first follow the pipelines proposed by \cite{wu2020unsupervised,pan20202d}, where we manipulate the StyleGAN images to produce images with various viewpoints and lighting conditions.
As indicated in Figure \ref{fig:3d}, we take the auto-encoding architecture. 

Specifically, for a given image $\mathbf{I}$, we aim to predict its depth map $\bm{d}$, an albedo image $\bm{a}$, a viewpoint $\bm{v}$, and a light direction $\bm{l}$. These four factors can be predicted from the VLDA module $(V, L, D, A)$, which are constructed by different learn-able networks. With the above four factors, we can reconstruct the input images $\mathbf{\hat{I}}$ with the rendering process $\Phi$:
\begin{align}
\mathbf{\hat{I}} &=\Phi(\bm{d}, \bm{a}, \bm{v}, \bm{l}).
\end{align}

In the first step, we initialize the shape prior with an ellipsoid shape, and use the canonical setting for the lighting and viewpoint conditions. We produce $\mathbf{\hat{I}}_1$ with the albedo network $A$ and use the reconstruction loss between $\mathbf{\hat{I}}_1$ and $\mathbf{I}$ to train $A$, which are formulated as
\begin{align}
\Loss_{recon}(\mathbf{\hat{I}}_1, \mathbf{I}) = ||\mathbf{\hat{I}}_1 - \mathbf{I}|| + \lambda_{perc} ||F(\mathbf{\hat{I}}_1) - F(\mathbf{I})||^2_2,
\end{align}
where the former and latter terms represent the L1 loss and the perceptual loss \cite{johnson2016perceptual} respectively. $F(\cdot)$ denotes the VGG \cite{simonyan2014very} feature extraction model. 

However, we observe if we directly use $\Loss_{recon}$ as \cite{pan20202d} above, the side views of some avatars would have distortions and the two sides of the cartoon face may present obvious discrepancy. This is because of the limited capability of StyleGAN model inferring the novel views based on the single-view images.
Hence, we propose to give an extra symmetry loss to regularize the rendering process, which can be denoted as 
\begin{align}
\Loss_{sym}(\mathbf{\hat{I}}_1^{f}, \mathbf{I}) = ||\mathbf{\hat{I}}_1^{f} - \mathbf{I}|| + \lambda_{perc} ||F(\mathbf{\hat{I}}_1^{f}) - F(\mathbf{I})||^2_2.
\end{align}
$\mathbf{\hat{I}}_1^{f}$ depicts the flipped reconstructed images. Using $\Loss_{sym}$ allows the model to refine the distorted face regions based on the symmetrical parts, which resolves the issue of side-view distortions to some extent.

Moreover, to ensure the identity consistency during the 3D reconstruction process, we also apply the face ID loss on the reconstructed and input images. We adopt the same pre-trained face recognition network as Eq. \ref{eq:id}. For simplicity, we denote the ID loss as $\Loss_{id}(\mathbf{\hat{I}}_1, \mathbf{I})$. The overall learning objective of the first step is formulated as:
\begin{align}
\Loss_1 = \Loss_{recon}(\mathbf{\hat{I}}_1, \mathbf{I}) + \Loss_{sym}(\mathbf{\hat{I}}_1^f, \mathbf{I}) + \Loss_{id}(\mathbf{\hat{I}}_1, \mathbf{I}).
\end{align}

In the second step, we randomly sample various viewpoints $\{\bm{v}_i | i = 1,2,...,m\}$ and lighting conditions $\{\bm{l}_i| i = 1,2,...,m\}$, which are used together with the depth $\bm{d}_0$ and albedo $\bm{a}_0$ that are computed at the first step to generate the rendered images. However, the rendered images may fail to present good results, since the estimated $\bm{d}_0$ is not accurate enough at the initial training phase.
Hence, we then project the rendered images back to the latent space $\W$ of StyleGAN, which gives strong regularization on the projected images, such that they can have better quality.
Note that here we have the latent codes $\w$ for the original input images as discussed in Section \ref{sec:latent}, hence we predict the offset $\Delta \w$ with an encoder $E(\cdot)$ to ease the training difficulty \cite{pan20202d}, which can be formulated as:
\begin{align}
\Delta \w = & E(\mathbf{I}). \\
\mathbf{I}_{p} = & G\big(\w + E(\mathbf{I})\big).
\end{align}
Here $\mathbf{I}_{p}$ denotes the projected sample.
The whole optimization process can be denoted as
\begin{align} \label{eq:2}
\Loss_2 = & \Loss_{dis}(\mathbf{I}_{p}, \mathbf{I}) + \lambda_E \| \Delta \w \|_2.
\end{align}
$\Loss_{dis}$ denotes the L1 distance of discriminator features, which are demonstrated by \cite{pan2021exploiting} to be more useful at the generated samples. The latter term of Eq. \ref{eq:2} is used to avoid the learned offset being too large.

In the third step, we aim to jointly train the $(V, L, D, A)$ networks. We denote the reconstructed images as:
\begin{align}
\hat{\mathbf{I}}_{3} = \Phi\big(D(\mathbf{I}), A(\mathbf{I}), V(\mathbf{I}_{p}), L(\mathbf{I}_{p})\big).
\end{align}
The learning objective of the third step is:
\begin{align}
\Loss_3 = &\Loss_{recon}(\hat{\mathbf{I}}_{3}, {\mathbf{I}}_{p}) + \Loss_{sym}(\mathbf{\hat{I}}_3^{f}, {\mathbf{I}}_{p}) + \Loss_{id}(\mathbf{\hat{I}}_3, \mathbf{I})\nonumber \\
&  + \lambda_{smooth} \Loss_{smooth} \big(D(\mathbf{I})\big).
\end{align}
$\Loss_{smooth}$ is defined in \cite{zhou2017unsupervised}, which minimizes L1 norm of the second-order gradients for the predicted depth maps. Training one cycle of the above three steps is not enough to reconstruct the 3D shape with fine details, hence we repeat these three steps four times to refine the 3D reconstructed results.  

\section{Experiments}
\label{sec:exp}

\subsection{Setup}
\noindent \textbf{Datasets and metrics.}
We test our method on three cartoon face datasets: Disney \cite{pinkney2020resolution}, MetFaces \cite{karras2020training} and Ukiyo-e \cite{pinkney2020resolution}. Disney, MetFaces and Ukiyo-e contain $317$, $1336$ and $5209$ images respectively. We evaluate the quality and diversity of the rendered 2D images from the 3D shapes with the Fr\'echet Inception Distance (FID) \cite{heusel2017gans} and perceptual loss $\Loss_{perc}$ \cite{johnson2016perceptual}. 
As these datasets do not have any ground-truth annotations for the 3D shapes, we use the BFM dataset \cite{paysan20093d} to evaluate the 3D reconstruction results.

\begin{figure}
\begin{center}
\includegraphics[width=0.48\textwidth]{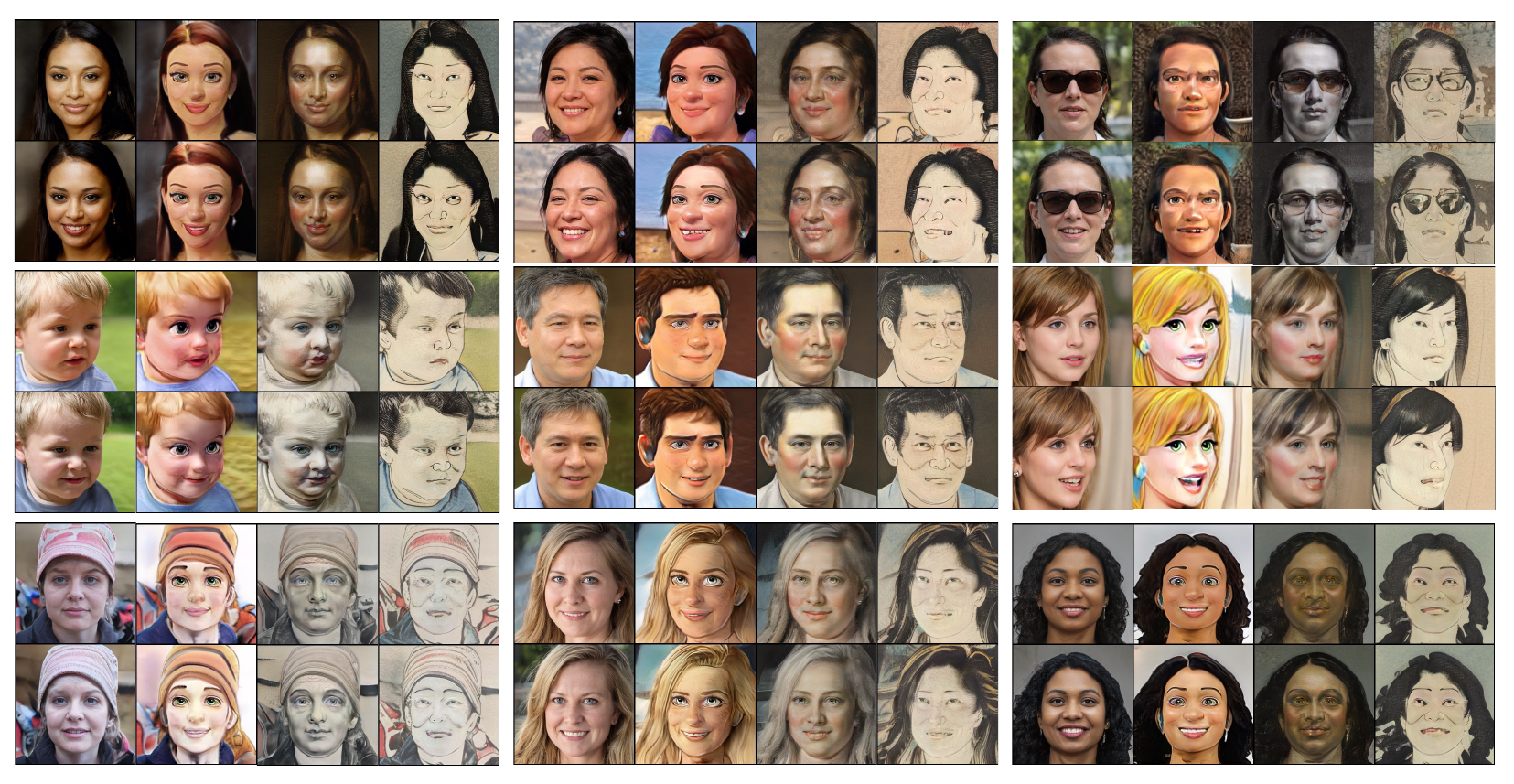}
\vspace{-10pt}
\end{center}
   \caption{Visualization of the manipulated facial expressions. In each block, from left to right, we show consistent results of natural human face images, Disney-style, Metfaces-style and Ukiyoe-style images respectively. }
\vspace{-10pt}
\label{fig:ex6}
\end{figure}

\noindent \textbf{Implementation details.}
We use the StyleGAN2 \cite{Karras2019stylegan2} model to train the cartoon datasets, which is based on the PyTorch implementation\footnote{https://github.com/rosinality/stylegan2-pytorch}. We apply the structure loss on the output of the first $k=2$ blocks. For all of our used datasets, we resize the original images to the resolution of $256$ and train the cartoon StyleGAN model. 
In the latent code manipulation process, we set the initial learning rate as $0.1$ and the learning rate gradually decreases, which is the same strategy as used in \cite{Karras2019stylegan2}. For each sample, we optimize it for $10$ iterations to give the $w^*$. We adopt the pretrained face embedding model used in \cite{shi2021lifting} for $\Loss_{id}$ and $\Loss_{low}$, which is an ResNet-18 \cite{he2016deep}. 
The structures of $V, L, D, A$ networks are the same as \cite{wu2020unsupervised}. Specifically, the depth and albedo networks are built by the encoder-decoder architecture, while viewpoint and lighting networks are implemented with encoders to give regressed results. We follow \cite{pan20202d} to implement the GAN inversion encoder $E$, which uses the ResNet \cite{he2016deep} architecture. The hyperparamters $\lambda_{low}$, $\lambda_{perc}$, $\lambda_{E}$ and $\lambda_{smooth}$ are set $0.2$, $0.5$, $0.01$ and $0.01$ respectively. The hyperparamters are chosen based on the empirical observations during model training.

\subsection{Facial expression manipulation results}
In Figure \ref{fig:ex6}, we show the facial expression manipulation results across three different datasets, i.e. Disney \cite{pinkney2020resolution}, MetFaces \cite{karras2020training} and Ukiyo-e \cite{pinkney2020resolution}. We can see the overall consistency between the generated cartoon faces and the original human faces. To be specific, we first randomly sample latent codes of StyleGAN, where we further discover their semantic directions and present images with different facial expressions. The latent codes that we feed into the cartoon StyleGAN model are the same as that in the human face model, hence the cartoon faces can share similar semantic concepts as the original human faces. For example, in the top-right block of Figure \ref{fig:ex6}, our generated MetFaces-style and Ukiyoe-style cartoon faces have sunglasses, which do not exist in the original datasets but are consistent with the given human faces.  
Here we use SeFa \cite{shen2021closed} to find the initial offsets to manipulate the latent codes, then we propose to use the face identity loss and the low-level feature regularization loss to optimize the initial offsets.

\begin{table}
  \centering
    \caption{Quantitative comparisons on FID and perceptual loss $\Loss_{perc}$ evaluated on the three datasets. We also give ablation results to analyze the usefulness of structure loss and model interpolation.}
  \begin{center}
  \scalebox{0.8}{
  \begin{tabular}{l|cccccc}
\toprule
\textbf{Method}    & \multicolumn{2}{c}{\textbf{Disney} \cite{pinkney2020resolution}}  & \multicolumn{2}{c}{\textbf{MetFaces} \cite{karras2020training}}   & \multicolumn{2}{c}{\textbf{Ukiyo-e} \cite{pinkney2020resolution}}      \\
\textbf{Metrics} & FID & $\Loss_{perc}$ & FID & $\Loss_{perc}$ & FID & $\Loss_{perc}$  \\
\midrule
Unsup3d \cite{wu2020unsupervised}    & 205.6 & 0.71 & 192.5 & 0.71 & 316.8 & 0.73  \\
LiftedGAN \cite{shi2021lifting} & 185.2  & 0.67 & 166.6 & 0.65 & 284.8 & 0.76    \\
\midrule
Ours   & \textbf{146.8}  & \textbf{0.65} & \textbf{140.7} & \textbf{0.64} & \textbf{188.1} & \textbf{0.71} \\
\quad - w/o structure loss   &  159.9 & 0.65  & 156.9  & 0.70 & 280.3 & 0.75  \\
\quad - w/o model interpolation   & 181.7 & 0.70 & 189.5 & 0.64 & 234.0 & 0.72 \\
\bottomrule
  \end{tabular}
  }
  \vspace{-10pt}
  \end{center}
  \label{tab:result}
\end{table}

\begin{table}
\centering
\caption{Quantitative evaluation on the 3D Reconstruction process. We adopt the scale-invariant depth error (SIDE) and mean angle deviation (MAD) scores as the metrics. $^\dag$ indicates Unsup3d output resolutioin is 64. $^*$ indicates we report our reproduced results for GAN2Shape.}\label{tab:3d}
\begin{tabular}{l|cc}\toprule
\textbf{Method} & \textbf{SIDE ($\times10^{-2}$)} $\downarrow$  & \textbf{MAD} $\downarrow$ \\\midrule
Unsup3d$^\dag$ \cite{wu2020unsupervised} &0.793 &16.51  \\
GAN2Shape$^*$ \cite{pan20202d} & 0.838 & 15.80 \\
\midrule
Ours& \textbf{0.802} & \textbf{15.41} \\
\quad - w/o $\Loss_{id}$ & 0.846 & 15.92 \\
\quad  - w/o $\Loss_{sym}$ & 0.881 & 16.53 \\
\bottomrule
\end{tabular}
\vspace{-10pt}
\end{table}

\subsection{Comparison with other works}
\subsubsection{Quantitative evaluation on generated cartoon faces}
In Table \ref{tab:result}, we show the quantitative results of Unsup3d \cite{wu2020unsupervised}, LiftedGAN \cite{shi2021lifting} and our proposed method. We also give ablative studies regarding the structure loss and model interpolation. For each dataset we sample $256$ rendered 2D images for metrics calculation, and we use Fr\'echet Inception Distance (FID) \cite{heusel2017gans} and the perceptual loss \cite{johnson2016perceptual} $\Loss_{perc}$ to do the evaluation. To be specific, FID is to measure the distribution similarities between the rendered cartoon faces and the original FFHQ \cite{karras2019style} datasets, and $\Loss_{perc}$ is to measure the instance similarity between cartoon and human faces. We observe our method performs better than Unsup3d \cite{wu2020unsupervised} and LiftedGAN \cite{shi2021lifting}. Specifically, model interpolation improves our results most, since it helps generate more photo-realistic cartoon images.

\subsubsection{Quantitative evaluation for 3D reconstruction}
In Table \ref{tab:3d}, we present the quantitative evaluation the 3D reconstruction process. Specifically, we adopt the BFM dataset \cite{paysan20093d}, which is a synthetic face model having the ground truth human face 3D annotations. We use the same metrics as  \cite{wu2020unsupervised} and \cite{pan20202d}, i.e. scale-invariant depth error (SIDE) \cite{eigen2014depth} and mean angle deviation (MAD). These metrics evaluate the accuracy of the predicted depth maps. We observe with the adopted symmetry and person identity losses, the reconstructed geometry yields better performance, as they provide stronger regularization. Specifically, the symmetry loss $\Loss_{sym}$ is more useful in terms of the 3D reconstruction, which utilizes the inherent characteristics of human faces to reduce the reconstruction difficulty. 

\begin{figure}
\begin{center}
\includegraphics[width=0.48\textwidth]{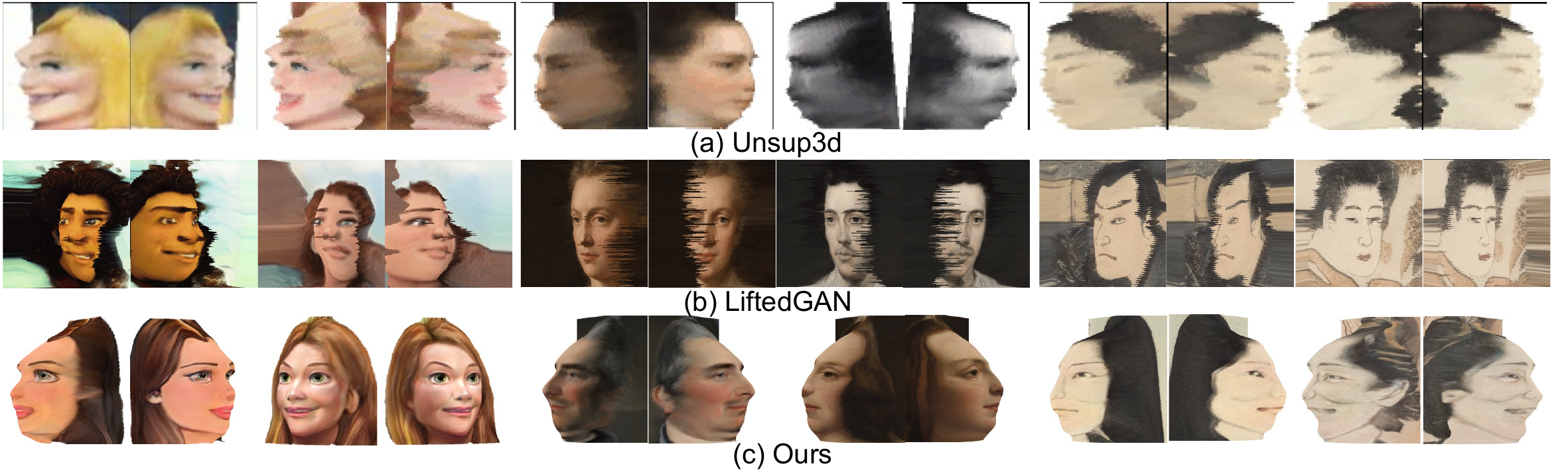}
\end{center}
\vspace{-10pt}
  \caption{Comparisons of qualitative results between ours and two related works: (a) Unsup3d \cite{wu2020unsupervised} and (b) LiftedGAN \cite{shi2021lifting}.}
\vspace{-10pt}
\label{fig:ex4}
\end{figure}

\begin{figure*}
\begin{center}
\includegraphics[width=0.75\textwidth]{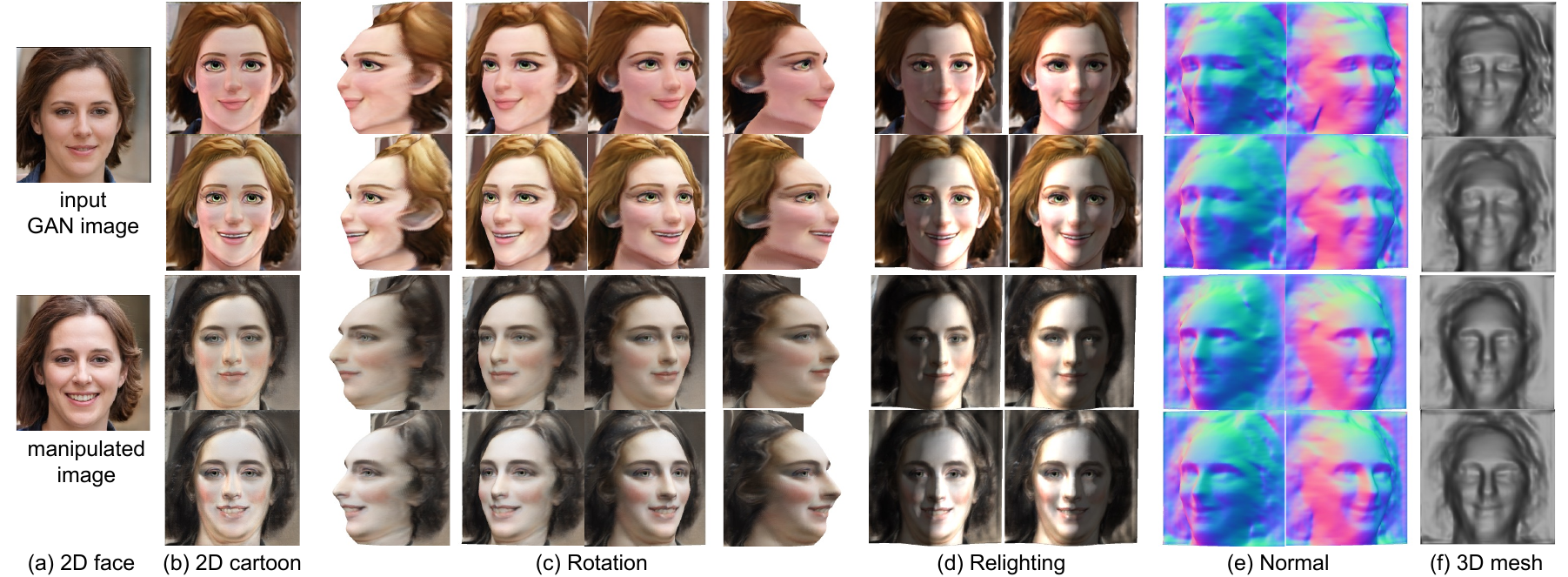}
\end{center}
\vspace{-10pt}
  \caption{In Column (a), input a GAN image, we can manipulate the input face image by changing the StyleGAN latent code. The corresponding generated cartoon styled images are presented Column (b). Column (c) and (d) present the rendered results of various viewpoints and lighting conditions. Column (e) and (f) show the normal maps and 3D mesh.}
\label{fig:teaser}
\end{figure*}

\begin{figure*}
\begin{center}
\includegraphics[width=0.7\textwidth]{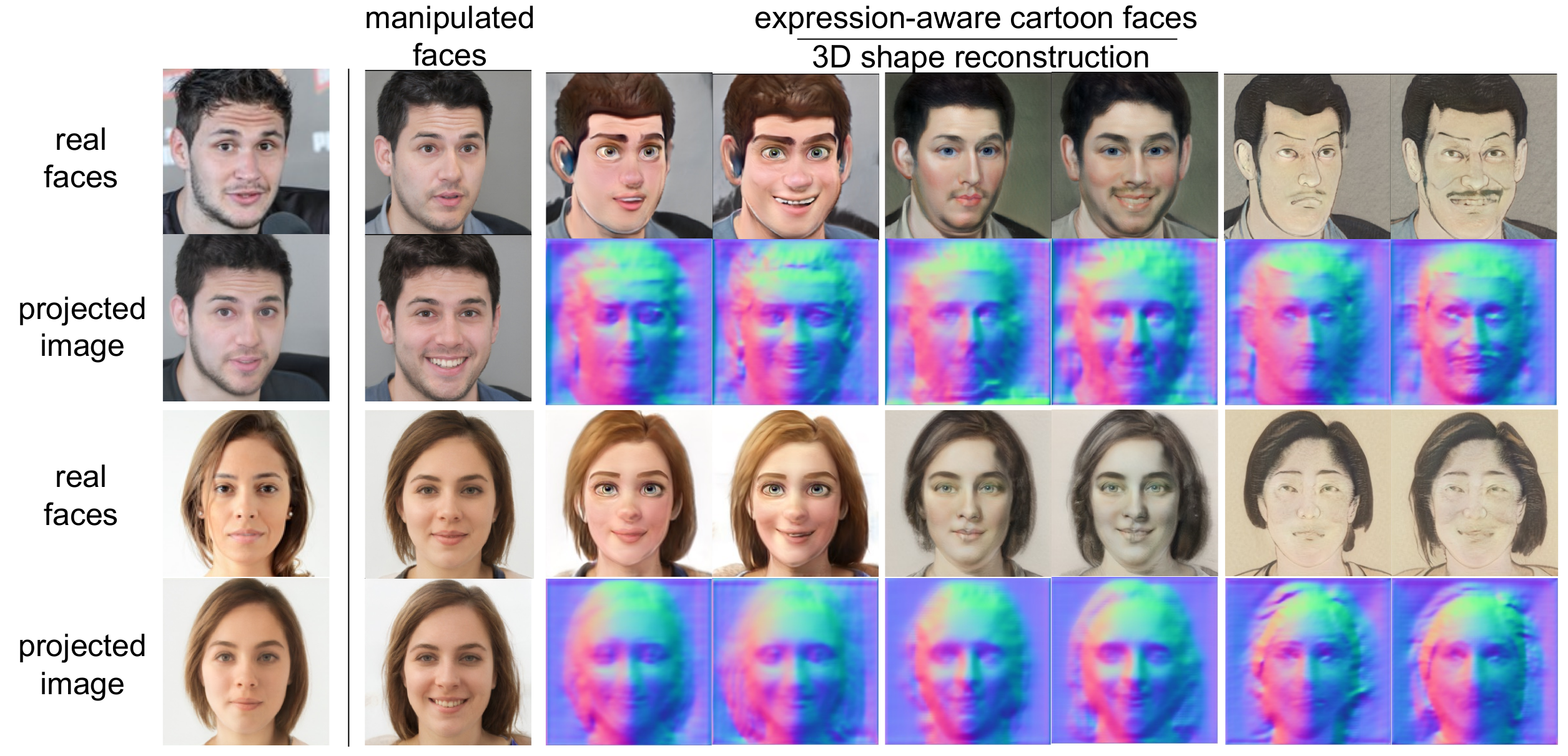}
\vspace{-15pt}
\end{center}
   \caption{Generated results from real face images. In the first column, we show the real faces and reconstructed images generated by the projected latent codes. In the second column, we show the faces with different facial expressions based on the manipulated latent codes. In the following columns, we show the cartoon faces and their reconstructed 3D shapes.}
\vspace{-10pt}
\label{fig:ex1}
\end{figure*}

\subsubsection{Qualitative results}
In Figure \ref{fig:ex4}, we show the rotated results of related works \cite{wu2020unsupervised,shi2021lifting} using the unsupervised method to reconstruct the 3D shapes. To be specific, we observe the 3D shapes reconstructed by Unsup3d \cite{wu2020unsupervised} have unnatural distortions. Besides, as the rendered image size of Unsup3d is set as $64$, the results also lack fine-grained details. Both LiftedGAN \cite{shi2021lifting} and our method set the resolution as $256$. LiftedGAN \cite{shi2021lifting} performs well at generating the front-view cartoon faces only, while at other viewpoints, it renders low-quality images with large distortions, since LiftedGAN simply infers the depth from latent codes and trains the model without any constraint on the depth. In contrast, we initialize the shape prior with an ellipsoid shape and refine the reconstructed 3D shapes with multiple cycles, hence we can obtain results with better quality.

\subsection{3D shape reconstruction}

In Figure \ref{fig:teaser}, we show the visualizations from the 2D human faces to the learned 3D shapes. 
The shown rotation and relighting results are rendered from the reconstructed 3D shapes. Generally, our framework captures the fine-grained details of the 2D images, e.g. the face wrinkles and skin color. 
The surface normal maps are used to generate face images with arbitrary lighting conditions. The relighting samples preserve the original face information with only lighting changed, indicating the good quality of our learned normal maps. 

\subsection{Applications on real human faces}
In our previous experiments, we directly sample random latent codes $\w$ to generate human faces and cartoon avatars.
If one wants to experiment with the real face data, existing GAN inversion methods \cite{Karras2019stylegan2,zhu2016generative,bau2020semantic,zhu2020indomain} can be used to project the real face images back to the latent space $\mathcal{W}$ and obtain the latent codes $\rm\bf w$. This means in GAN inversion, we aim to find $\w$ that can reconstruct the given image with the trained GAN. Here we directly take the image projection method given by the StyleGAN2 implementation\footnote{https://github.com/rosinality/stylegan2-pytorch}, we show the qualitative results in Figure \ref{fig:ex1}. It is notable that better GAN inversion results can be expected if more advanced methods are adopted. Generally, we observe our method gives good manipulation and 3D reconstruction results.

\begin{figure}
\begin{center}
\includegraphics[width=0.48\textwidth]{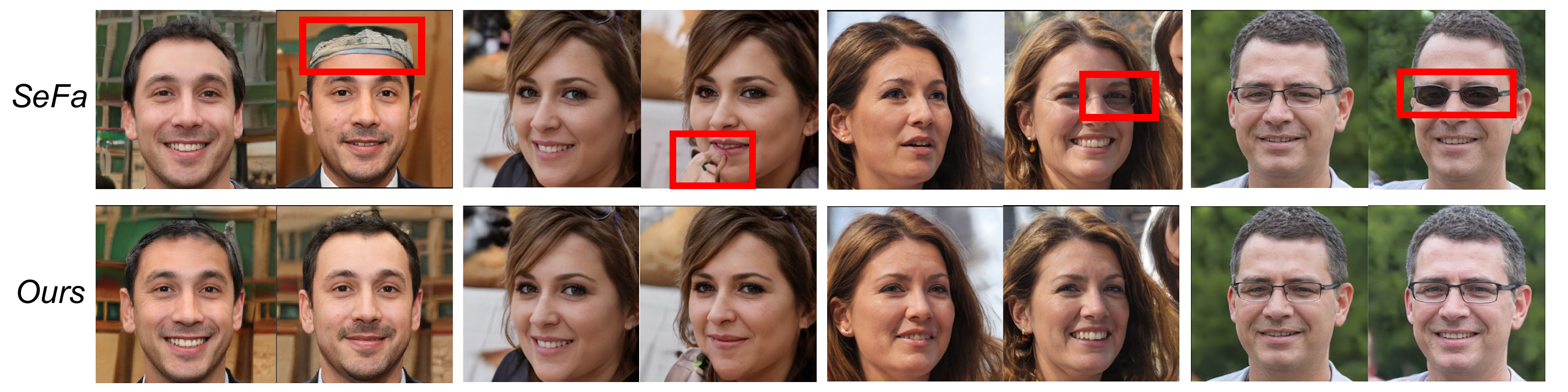}
\end{center}
\vspace{-5pt}
   \caption{Comparisons between SeFa \cite{shen2021closed} and our proposed latent code manipulation method. In each column, we show results generated by the manipulated latent codes, i.e. $\w + \Delta \w$ and $\w - \Delta \w$. The upper and bottom columns show images generated by SeFa \cite{shen2021closed} and our proposed method respectively. The red boxes depict the unwanted manipulated parts by SeFa \cite{shen2021closed}, including the \emph{hat}, the \emph{distorted face shape}, the \emph{half eyeglasses}, and the \emph{changed eyeglass colour}. While our method properly alleviates these issues. }
   \vspace{-10pt}
\label{fig:ex2}
\end{figure}

\begin{figure}
\begin{center}
\includegraphics[width=0.4\textwidth]{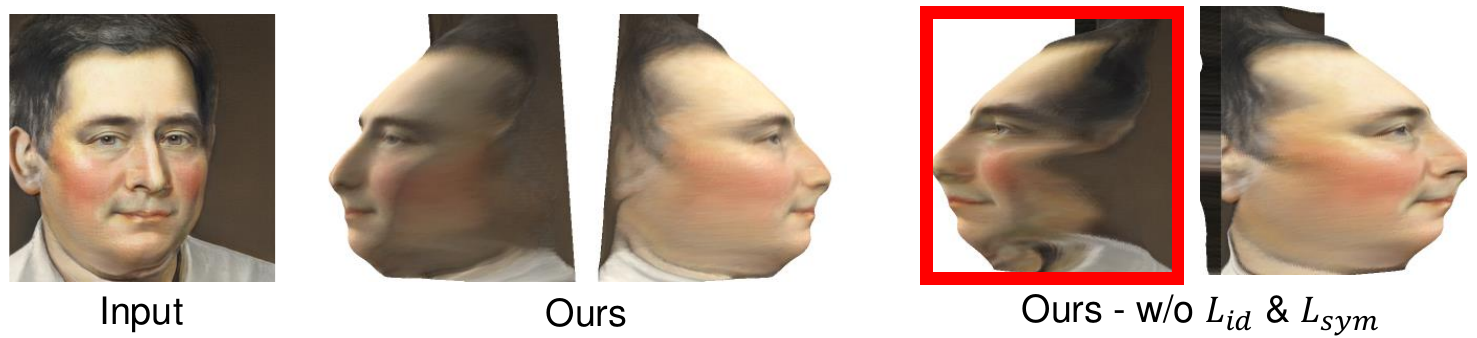}
\end{center}
\vspace{-10pt}
  \caption{The result visualization of the input images, our proposed method, and our method without $\Loss_{id}$ and $\Loss_{sym}$. The red boxes indicate the unnatural or distorted areas from the ablative results.}
\vspace{-10pt}
\label{fig:flip}
\end{figure}

\subsection{Ablation studies}
\noindent \textbf{Efficacy of the latent code optimization.}
In Figure \ref{fig:ex2}, we present the qualitative results of our proposed latent code manipulation method. We first follow the state-of-the-art method SeFa \cite{shen2021closed} with the closed-form factorization to find the semantic directions of the given latent codes $\w$, then we present images generated from the manipulated latent codes: $\w + \Delta \w$ and $\w - \Delta \w$. As shown in Figure \ref{fig:ex2}, directly taking the $\Delta \w$ of SeFa \cite{shen2021closed} can change the facial expressions, while it fails to make other semantic concepts unchanged. For example, in the first column, the SeFa \cite{shen2021closed} manipulated image appears to have an unwanted bands around the head. In contrast, our results not only keep the original face identity, but also provide apparent facial expression variations.

\noindent \textbf{Efficacy of the person identity loss and symmetry loss during the 3D reconstruction process.}
In Figure \ref{fig:flip}, we qualitatively present the usefulness our adopted person identity loss $\Loss_{id}$ and symmetry loss $\Loss_{sym}$, in which we compare the results with and without $\Loss_{id}$ and $\Loss_{sym}$. 
If we remove $\Loss_{id}$ and $\Loss_{sym}$, we observe the reconstructed side-view images of some avatars may have distortions. 
While training with $\Loss_{id}$ and $\Loss_{sym}$ yields more natural reconstructed results.

\section{Conclusions}
This paper presents a novel framework for the expression-aware 3D cartoon face shape generation from a single GAN generated human face image. Specifically, 
we propose to optimize the manipulated offsets of the latent codes to make only one semantic attribute changed and preserve other attributes. To stylize the given human face images, we utilize transfer learning to train a cartoon face generation model, given limited cartoon data.
During the 3D shape reconstruction process, we manipulate the latent codes to give pseudo samples of different viewpoints and lighting conditions to enable 3D learning.
We demonstrate the efficacy of our proposed framework in the 3D cartoon face generation and manipulation. 

\section*{Acknowledgment}
This research is supported by the RIE2025 Industry Alignment Fund – Industry Collaboration Projects (IAF-ICP) (Award I2301E0026), administered by A*STAR, as well as supported by Alibaba Group and NTU Singapore through Alibaba-NTU Global eSustainability CorpLab (ANGEL).

\bibliographystyle{IEEEtran}
\bibliography{egbib}

\end{document}